\documentclass{article}
\usepackage{ijcai16}
\usepackage{times}
\usepackage{url}
\usepackage{latexsym} 
\usepackage{graphicx}
\usepackage{amscd, amsmath, amsthm, amssymb}
\usepackage{bbm}
\usepackage{mathtools}
\usepackage{booktabs,dcolumn}
\usepackage{algorithm, algorithmic}
\usepackage{caption}
\usepackage{subcaption}
\usepackage{multicol}
\usepackage{etoolbox}
\apptocmd{\sloppy}{\hbadness 10000\relax}{}{}

\newtheorem{definition}{Definition}

\newtheorem{theorem}{Theorem}
\newtheorem{property}{Property}

\newtheorem{remark}[theorem]{Remark}

\title{Exploiting Problem Structure in Combinatorial Landscapes: \\A Case Study on Pure Mathematics Application}

\author{Xiao-Feng Xie\\ 
WIOMAX LLC, USA\\
xie@wiomax.com
\And
Zun-Jing Wang\\ 
WIOMAX LLC, USA\\
wang@wiomax.com
}

\begin{document}

\maketitle

\begin{abstract}
In this paper, we present a method using AI techniques to solve a case of pure mathematics applications for finding narrow admissible tuples. The original problem is formulated into a combinatorial optimization problem. In particular, we show how to exploit the local search structure to formulate the problem landscape for dramatic reductions in search space and for non-trivial elimination in search barriers, and then to realize intelligent search strategies for effectively escaping from local minima. Experimental results demonstrate that the proposed method is able to efficiently find best known solutions. This research sheds light on exploiting the local problem structure for an efficient search in combinatorial landscapes as an application of AI to a new problem domain.
\end{abstract}

\section{Introduction}

AI techniques have shown their advantages on solving different combinatorial optimization problems, such as satisfiability \cite{sutton2010directed,dubois2001backbone,bjorner2015maximum_s,ansotegui2015exploiting_s}, traveling salesman problem \cite{zhang2004phase}, graph coloring \cite{culberson2001frozen}, job shop scheduling  \cite{watson2003problem}, and automated planning \cite{bonet2001planning}.

These problems can be generalized into the concept of combinatorial landscapes \cite{reidys2002combinatorial,schiavinotto2007review,tayarani2014landscape}, and problem solving can be cast to a search over a space of states. Such problems often are very hard \cite{cheeseman1991really}. In order to pursue an efficient search, it is vital to develop techniques to identify problem features and of exploiting the local search structure \cite{frank1997gravity,hoffmann2001local,hoos2004stochastic}. In particular, it is important to reduce and decompose the problem preserving structural features that enable heuristic search and with the effective problem size factored out \cite{slaney2001backbones,mears2015towards_s}. It is also significant to tackle ruggedness \cite{billinger2013search} and neutrality (plateaus) \cite{collins2006finding,benton2010g} in the landscapes.   

In number theory, a $k$-tuple $\mathcal{H}_k$ is {\em admissible} if $\phi_p (\mathcal{H}_k) < p$ for every prime $p$, where $\mathcal{H}_k=(h_1,\ldots,h_k)$ is a strictly increasing sequence of integers, and $\phi_p (\mathcal{H})$ denotes the number of distinct residue classes modulo $p$ occupied by the elements in $\mathcal{H}_k$ \cite{goldston2009primes}. 
The objective is to minimize the \emph{diameter} of~$\mathcal{H}_k$, i.e., $d(\mathcal{H}_k)=h_k-h_1$, for a given $k$. 

The early work \cite{hensley1974primes,gordon1998dense,clark2001dense} to compute narrow admissible tuples has been motivated by the incompatibility of the two long-standing Hardy-Littlewood conjectures.

Admissible sets have been used in the recent breakthrough work to find small gaps between primes. 
In \cite{goldston2009primes}, it was proved that any admissible $\mathcal{H}_{k_0}$ contains at least two primes infinitely often, if $k_0$ satisfies some arithmetic properties. 
In \cite{zhang2014bounded}, it was proved that a finite bound holds at $k_0 \geq 3.5 \times 10^6$. The bound was then quickly reduced to  $d^*(\mathcal{H}_{105})=600$ \cite{maynard2015small} and $d^*(\mathcal{H}_{50})=246$ \cite{polymath2014variants}, and wider ranges of $k_0$ also were obtained on bounded intervals containing many primes \cite{polymath2014variants}. Moreover, admissible sets have been used to find large gaps between primes \cite{ford2015chains}.

Most of the existing techniques to find narrow admissible tuples are {\em sieve} methods \cite{hensley1974primes,clark2001dense,gordon1998dense,polymath2014as,polymath2014variants}, although a few local optimizations were proposed recently \cite{polymath2014as,polymath2014variants}.   

In this paper, we formally model this problem into a combinatorial optimization problem, and design search strategies to tackle the landscape, by utilizing the local search structure. Our solver is systematically tested to show its effectiveness.

\section{Search Problem Formulation} \label{sec:model}

For a given $k$, a {\em candidate number set} $\mathcal{V}$ with $|\mathcal{V}| \ge k$ can be precomputed, and a required {\em prime set} $\mathcal{P}$, where each prime $p \le k$, can be determined. Each $\mathcal{H}$ is obtained by selecting the numbers from $\mathcal{V}$, and the admissibility is tested using $\mathcal{P}$.  

\begin{definition}[Constraint Optimization Model]
\label{def:COM} For a given $k$, and given the required $\mathcal{V}$ and $\mathcal{P}$, the objective is to find a number set $\mathcal{H} \subseteq \mathcal{V}$ with the minimal $d(\mathcal{H})$ value, subject to the constraints $|\mathcal{H}|=k$ and $\mathcal{H}$ is admissible. 
\end{definition}

For convenience, $\mathcal{V}$, $\mathcal{P}$, and $\mathcal{H}$ are assumed to be sorted in increasing order. We denote $\mathcal{H}$ as $\mathcal{H}_k$ if $|\mathcal{H}|=k$, as $\tilde{\mathcal{H}}$ if it is {\em admissible}, and as $\tilde{\mathcal{H}}_k$ if it satisfies both of the constraints. 

Given $\mathcal{V}$ and $\mathcal{P}=(p_1, \cdots, p_i, \cdots, p_{|\mathcal{P}|})$, the following three data structures are defined for facilitating the search:

\begin{definition}  [Residue Array $\mathcal{R}_v$]
\label{def:R} For $v \in \mathcal{V}$, $\mathcal{R}_v$ is calculated on $\mathcal{P}$: its $i$th row is $r_{v,i}=v \operatorname{mod} p_i$ for $i\in[1, |\mathcal{P}|]$. 
\end{definition}

\begin{definition} [Occupancy Matrix $\mathcal{M}$] 
\label{def:M} $\mathcal{M}$ is an irregular matrix, in which each row $i$ contains $p_i$ elements corresponding to the residue classes modulo $p_i$. For any given number set $\mathcal{H} \subseteq \mathcal{V}$, there is $m_{i,j}=\sum_{v\in\mathcal{H}} \mathbbm{1}(j \equiv r_{v,i}+1)$ for $j\in[1, p_i]$, which means the count of numbers in $\mathcal{H}$ occupying each residue class modulo $p_i$. 
\end{definition}

\begin{definition} [Count Array $\mathcal{F}$]
\label{def:F} $\mathcal{F}$ is an array, in which each row $i$ gives the count of zero elements in the $i$th row of $\mathcal{M}$. 
\end{definition}

\begin{property}
\label{thm:spacereq}  
The space requirements for $\{\mathcal{R}_v | v\in \mathcal{V}\}$, $\mathcal{M}$, $\mathcal{F}$ are respectively $|\mathcal{V}|\cdot|\mathcal{P}|$, $\sum_{i\in[1,|\mathcal{P}|]} p_i$, and $|\mathcal{P}|$.
\end{property}

\begin{figure} [ht]
\centering \includegraphics[width=0.42\textwidth]{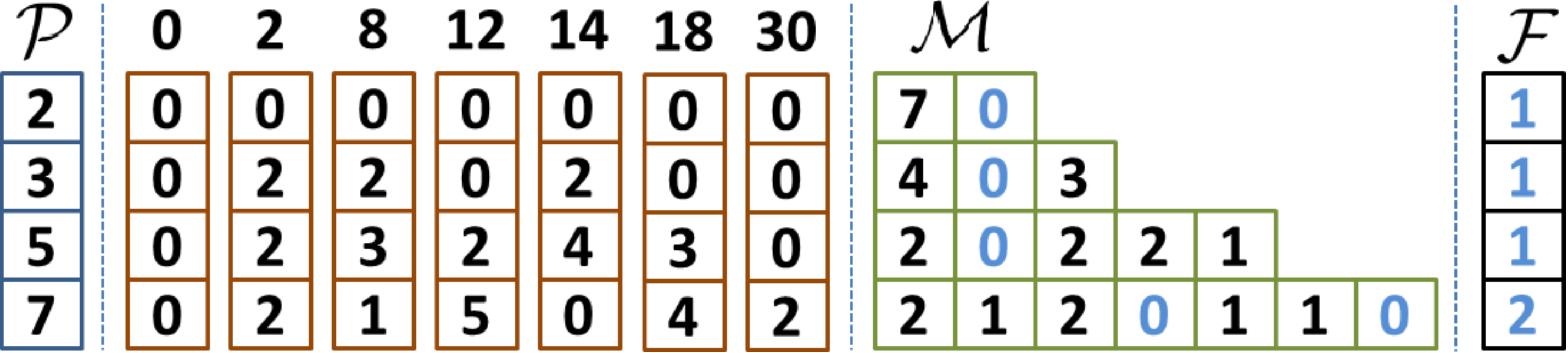} \caption{An admissible example for $\tilde{\mathcal{H}}_7$ with $d(\tilde{\mathcal{H}}_7)=30$.}
\label{fig:admissibleinst} 
\end{figure}

Figure \ref{fig:admissibleinst} gives the example for an admissible set $\tilde{\mathcal{H}}_7=(0, 2, 8, 12, 14, 18, 30)$ with $d(\tilde{\mathcal{H}}_7)=30$, the full prime set $\mathcal{P}$,  $\mathcal{R}_v$ for each $v \in \tilde{\mathcal{H}}_7$, and the corresponding $\mathcal{M}$ and $\mathcal{F}$. 

$\mathcal{H}$ and its corresponding $\mathcal{M}$ and $\mathcal{F}$ have a few properties:

\begin{property}[Admissibility]
\label{thm:AdmissibilityF}
$\mathcal{H}$ is admissible if $f_i>0$, $\forall i$.
\end{property}

There is a constraint violation at row $i$ if $f_i =p_i-\phi_{p_i} \equiv 0$. The total {\em violation count} should be 0 for each $\tilde{\mathcal{H}}$. 

\begin{property}
\label{thm:occuelems}  
Let $W_{i,j}=\{v\in \mathcal{H}|r_{v,i}\equiv j-1\}$, it contains all numbers occupying $(i,j)$ of $\mathcal{M}$, and there is $|W_{i,j}|=m_{i,j}$. 
\end{property}

\begin{property}
\label{thm:consevative}  
For each row $i$, there is $\sum_{j\in[1, p_i]}m_{i,j}=|\mathcal{H}|$.
\end{property}

There are two basic properties based on an admissible $\tilde{\mathcal{H}}_k$:
\begin{property}[Offsetting]
\label{thm:Offsetting}  
For any $c \in \mathbb{Z}$, $\mathcal{H}_k^{[c]}=(h_1+c, \ldots, h_k+c)$ is admissible, and there is  $d(\mathcal{H}_k^{[c]})=d(\tilde{\mathcal{H}}_k)$. 
\end{property}

\begin{property}[Subsetting]
\label{thm:Subsetting}  
Any subset of $\tilde{\mathcal{H}}$ is admissible. 
\end{property}

Properties \ref{thm:Offsetting} and \ref{thm:Subsetting} were observed in \cite{polymath2014variants}. Offsetting can be seen as rotating of residue classes at each row in $\mathcal{M}$; and subsetting does not decrease each row of $\mathcal{F}$.

Defining a compact $\mathcal{V}$ is nontrivial for reducing the size of problem space, which is exponential in $|\mathcal{V}|$. 

One plausible way is to let $\mathcal{V}$ include all numbers in $[0, U]$, and set $h_1=0$. Let $d_k^{LB}$ and $d_k^{UB}$ be the best-so-far lower and upper bounds of the optimal value of $d(\tilde{\mathcal{H}}_k)$. During the search, $|\mathcal{V}|=U$ can be bounded by $d_k^{UB}$. However, it appears that $d_k^{UB}$ is very close to $\lfloor k\log k + k \rfloor$ \cite{polymath2014variants}, which might still be very large when $k$ is big.

Based on Property \ref{thm:consevative}, as $p_i$ is small, the average $m_{i,j}$ would be large. For the rows with $f_i=1$, it would be difficult to find a useful heuristic for changing the unoccupied column $j$ at row $i$. Intuitively, each unoccupied location $(i,j)$ in $\mathcal{M}$ can be assumed to be unchanged during the search. 
Thus, {\em sieving} can be applied to remove any numbers in $[0, U]$ that occupy those unoccupied locations, which could be found using Property \ref{thm:occuelems}, for a set of the smallest primes $\mathcal{P}_R \in \mathcal{P}$. 
After sieving, the proportion of remaining numbers is $\alpha \approx 1-\prod_{p_i \in \mathcal{P}_R} (1-1/p_i)$. The completeness of the problem space on the other combinations 
in $\mathcal{M}$ can be retained using a sufficiently large $U$, based on the principle behind Property \ref{thm:Offsetting}, i.e., offsetting as a choice of residue classes. 

\begin{remark}
\label{thm:subproblem}
The original problem can be converted into a list of subproblems, where each subproblem takes each $v\in [0, U-d_k^{LB}] \cap \mathcal{V}$ as the starting point $h_1$ to obtain the minimal diameter for each $\tilde{\mathcal{H}}_k \subseteq  \mathcal{V}_v = [v, v+d_k^{UB}] \cap \mathcal{V}$. The optimal solution is then the best solution among all subproblems. 
\end{remark}

Decomposition \cite{friesen2015recursive_s} has been successfully used in AI for solving discrete problems. The new perspective of the problem has two features. First, each subproblem has a much smaller state space, as $|\mathcal{V}_v| \approx \alpha \cdot d_k^{UB}$. Second, for totally around $\alpha \cdot (U-d_k^{LB})$ subproblems,
the good solutions of neighboring subproblems might share a large proportion of elements, providing a very useful heuristic clue for efficient adaptive search in this dimension.

Let $\mathcal{P_C}$ contain all primes $p \le k$. In theory, $\mathcal{P}=\mathcal{P_C}$, but we can reduce it to an effective subset. Based on Property \ref{thm:consevative}, if $p_i$ is large, the average $m_{i,j}$ would be small. Some rows of $\mathcal{F}$ would always have $f_i>0$, even for $\mathcal{H} = \mathcal{V}$. The set of corresponding primes, named $\mathcal{P}_L$, thus can be removed from $\mathcal{P}$, without any loss of the completeness for testing the admissibility. The effective prime set would be $\mathcal{P}=\mathcal{P_C}-\mathcal{P}_L$. 

Algorithm \ref{alg:SieveVP} gives the specific realization for obtaining $\mathcal{V}$ and $\mathcal{P}$. Here $\mathcal{P}_R$ in Line 1 and $\mathcal{V}$ in Line 2 are obtained using the setting in the greedy sieving method \cite{polymath2014variants}. 

\begin{algorithm} [t]                    
\caption{Obtain $\mathcal{V}$ and $\mathcal{P}$}   
\label{alg:SieveVP}                           
\begin{algorithmic}[1]                    
\REQUIRE $k$, $[0, U]$ 
\STATE $\mathcal{P}_C=\{p \le k\}$; $\mathcal{P}_R=\{p<\sqrt{k\operatorname{log}k}\}$ \COMMENT{Let $p_{m}=\sqrt{k\operatorname{log}k}$}
\STATE $\mathcal{V}$ = $[0, U]$ sieving all $v$ with $r_{v,i} \equiv 1$ for $p_i \in \mathcal{P}_R$
\STATE Obtain $\mathcal{M}$ and $\mathcal{F}$ for $\mathcal{H}=\mathcal{V}$ using $\mathcal{P}=\mathcal{P}_C$
\STATE $\mathcal{P}_L = \{p_i \in \mathcal{P}_C| f_i>0\}$; $\mathcal{P}=\mathcal{P}_C-\mathcal{P}_L$ 
\RETURN $\mathcal{V}$,  $\mathcal{P}$
\end{algorithmic}
\end{algorithm}

\section{Search Algorithm}  \label{sec:algo}

In this section, the basic operations on auxiliary data structures are first introduced. Some search operators are then realized. Finally, we describe the overall search algorithm. 

\subsection{Operations on $\mathcal{R}_v$, $\mathcal{M}$ and $\mathcal{F}$}

For every $v \in \mathcal{V}$, $\mathcal{R}_v$ is calculated in advance. For $\mathcal{H} \subseteq \mathcal{V}$, the corresponding $\mathcal{M}$ and $\mathcal{F}$ are synchronously updated locally. The space requirements are given in Property \ref{thm:spacereq}.

There are two elemental 1-{\em move} operators, i.e., {\em adding} $v \notin \mathcal{H}$ into $\mathcal{H}$ to obtain $\mathcal{H}=\mathcal{H}+\{v\}$, and {\em removing} $v \in \mathcal{H}$ from $\mathcal{H}$ to obtain $\mathcal{H}=\mathcal{H}-\{v\}$, for given $\mathcal{H} \subseteq \mathcal{V}$ and $v \in \mathcal{V}$.

\begin{property}[Connectivity]
\label{thm:Connectivity}  
The two elemental 1-move operators possess the connectivity property for each $\mathcal{H} \in \mathcal{V}$.
\end{property}

The connectivity property \cite{nowicki1996fast} states that there exists a finite sequence of such moves to achieve the optimum state from any state in the search space. 

For $\mathcal{H}\equiv \varnothing$, there are $m_{i,j}=0$ for each $ i, j$, $f_i=p_i$ for each $i$, based on Definitions \ref{def:M} and \ref{def:F}. The $\mathcal{M}$ and $\mathcal{F}$ for any $\mathcal{H}$ can be constructed by adding each $v \in \mathcal{H}$ using Algorithm \ref{alg:AddingV}. For any two states $\mathcal{H}_A$ and $\mathcal{H}_B$, $\mathcal{H}_A$ can be changed into $\mathcal{H}_B$ by adding each $v \in \mathcal{H}_B - \mathcal{H}_A$ and by removing each $v \in \mathcal{H}_A - \mathcal{H}_B$. The total number of 1-moves is $L=|\mathcal{H}_A \cup \mathcal{H}_B|-|\mathcal{H}_A \cap \mathcal{H}_B|$, i.e., which can be seen as the {\em distance} \cite{reidys2002combinatorial} between two states. The shorter the distance, the more similar the two states are. 

Algorithms \ref{alg:AddingV} and \ref{alg:RemoveV} respectively give the operations of updating $\mathcal{M}$ and $\mathcal{F}$ for the two elemental 1-move operators.

\begin{algorithm} [t]                     
\caption{Update $\mathcal{M}$ and $\mathcal{F}$ as adding $v \notin \mathcal{H}$ into $\mathcal{H}$}   
\label{alg:AddingV}                          
\begin{algorithmic}[1]                    
\REQUIRE $R_v$, $\mathcal{M}$,  $\mathcal{F}$
\FOR{$i \in [1, |\mathcal{P}|]$}
\STATE $j=r_{v,i}+1$; $m_{i, j} = m_{i, j}+1$
\STATE {\bf if}~~$m_{i, j} \equiv 1$~~{\bf then}~~$f_i=f_i-1$
\ENDFOR
\RETURN $\mathcal{M}$,  $\mathcal{F}$
\end{algorithmic}
\end{algorithm}

\begin{algorithm} [t]                     
\caption{Update $\mathcal{M}$ and $\mathcal{F}$ as removing $v \in \mathcal{H}$ from $\mathcal{H}$}   
\label{alg:RemoveV}                          
\begin{algorithmic}[1]                    
\REQUIRE $R_v$, $\mathcal{M}$,  $\mathcal{F}$
\FOR{$i \in [1, |\mathcal{P}|]$}
\STATE $j=r_{v,i}+1$; $m_{i, j} = m_{i, j}-1$
\STATE {\bf if}~~$m_{i, j} \equiv 0$~~{\bf then}~~$f_i=f_i+1$
\ENDFOR
\RETURN $\mathcal{M}$,  $\mathcal{F}$
\end{algorithmic}
\end{algorithm}

\begin{algorithm} [t]                     
\caption{{\em  VioCheck}: Get the change of the violation count}  
\label{alg:VioCheck}                          
\begin{algorithmic}[1]                    
\REQUIRE $v$, $\mathcal{H}$ \COMMENT{Include $R_v$ and corresponding $\mathcal{M}$ and $\mathcal{F}$}
\STATE $\Delta=0$
\FOR{$i \in [1, |\mathcal{P}|]$}
\STATE {\bf if}~~$m_{i, r_{v,i}+1} \equiv 0$~~{\bf and}~~$f_i\equiv1$~~{\bf then}~~$\Delta=\Delta+1$
\ENDFOR
\RETURN $\Delta$ ~\COMMENT{The change of the violation count}
\end{algorithmic}
\end{algorithm}

\begin{property} [Time Complexity]
\label{thm:Complexity}   
Algorithms \ref{alg:AddingV} and \ref{alg:RemoveV} have the time complexity $O(|\mathcal{P}|)$ in updating $\mathcal{M}$ and $\mathcal{F}$. 
\end{property}

In the following realizations, we will focus on the search transitions between $\tilde{\mathcal{H}}$ states. The admissibility testing (Property \ref{thm:AdmissibilityF}) on each $\tilde{\mathcal{H}}$ is not explicitly applied. Instead, {\em VioCheck} in Algorithm \ref{alg:VioCheck} is used to check $\Delta$, i.e., the violation count to be increased, if adding $v$ into $\mathcal{H}$, using $R_v$, $\mathcal{M}$ and $\mathcal{F}$. 

\subsection{Search Operators}

We first realize some elemental and advanced search operators to provide the transitions between admissible states. 

\subsubsection{Side Operators}

Let $Side$=\{Left, Right\} define the two mutually reverse sides of ${\mathcal{H}}$. For an admissible state $\tilde{\mathcal{H}}$, each side operator tries to add or remove a number at the given side of $\tilde{\mathcal{H}}$ to obtain the admissible tuple with a diameter as narrow as possible.  

{\em SideRemove} just removes the element at the given $Side$ from $\tilde{\mathcal{H}}$, and its output is admissible, according to Property \ref{thm:Subsetting}. 

Algorithm \ref{alg:SideAdd} defines the operation {\em SideAdd} for adding a number $v$ into $\tilde{\mathcal{H}}$. To retain the admissibility, the number to be added is tested using Algorithm \ref{alg:VioCheck} to ensure the admissibility.

\begin{algorithm} [t]                     
\caption{{\em SideAdd}: For adding a number into $\tilde{\mathcal{H}}$}   
\label{alg:SideAdd}                           
\begin{algorithmic}[1]                   
\REQUIRE $\tilde{\mathcal{H}}$, $Side$ \COMMENT{Include \{$R_v$\} and corresponding $\mathcal{M}$ and $\mathcal{F}$}
\STATE {\bf if}~$Side \equiv$ Left~{\bf then}~~$v=h_1$~~{\bf else}~$v=h_{|\tilde{\mathcal{H}}|}$ \COMMENT{Get side value}
\STATE $l$ = {\em GetIndex}$(v, \mathcal{V})$ \COMMENT{Obtain the index $l$ of $v$ in $\mathcal{V}$}
\WHILE {$l \in [1, |\mathcal{V}|]$}
\STATE {\bf if}~~$Side \equiv$ Left~~{\bf then}~~$l=l-1$ ~~{\bf else}~~$l=l+1$ 
\STATE $\Delta=${\em VioCheck}($v_l$,  $\tilde{\mathcal{H}}$)  \COMMENT{Use Alg.~\ref{alg:VioCheck} to add the $l$th number in $\mathcal{V}$}
\STATE {\bf if}~~$\Delta \equiv 0$~~{\bf return}~$\tilde{\mathcal{H}}=\tilde{\mathcal{H}} \cup \{v_l\}$ ~\COMMENT{Ensure the admissibility}
\ENDWHILE
\RETURN $\tilde{\mathcal{H}}$ ~\COMMENT{The original $\mathcal{H}$ is unchanged}
\end{algorithmic}
\end{algorithm}

\subsubsection{Repair Operator}

The {\em Repair} operator repairs $\tilde{\mathcal{H}}$ into $\tilde{\mathcal{H}}_k$ using side operators: While $|\tilde{\mathcal{H}}|<k$, the {\em SideAdd} operator is iteratively applied on each side of $\tilde{\mathcal{H}}$, and the better one is kept; While $|\tilde{\mathcal{H}}|>k$, the {\em SideRemove} operator is iteratively applied on each side of $\tilde{\mathcal{H}}$, and the better one is kept. Finally, $\tilde{\mathcal{H}}_k$ is obtained as $|\tilde{\mathcal{H}}|=k$.

\subsubsection{Shift Search}

Algorithm \ref{alg:ShiftMove} gives the realization of the {\em ShiftSearch} operator. The side is selected at random (Line 2). Each shift \cite{polymath2014as} is realized by combining {\em SideRemove} and {\em SideAdd} (Line 4), leading to a distance of 2 to the original state. Starting from $\tilde{\mathcal{H}}_O$, we applies totally up to $N_L$ shifts (Line 3) unless {\em SideAdd} fails (Line 5), and the best state is kept as $\tilde{\mathcal{H}}_{N}$ (Line 6). The state $\tilde{\mathcal{H}}_{N}$ is accepted immediately if $d_{N} \le d_O$, or with a probability otherwise (Line 8), following the same principle as in simulated annealing \cite{kirkpatrick1983optimization}.

\begin{algorithm} [t]                     
\caption{{\em ShiftSearch}: Combine side moves on $\tilde{\mathcal{H}}$}   
\label{alg:ShiftMove}                          
\begin{algorithmic}[1]                    
\REQUIRE $\tilde{\mathcal{H}}_O$ ~\COMMENT{Parameter: $N_L \ge 1$, ${\beta} \ge 0$}
\STATE $\tilde{\mathcal{H}}=\tilde{\mathcal{H}}_{O}$; $k_O=|\tilde{\mathcal{H}}|$; $d_O=d(\tilde{\mathcal{H}})$; $d_{N}=\infty$; $\tilde{\mathcal{H}}_{N}=\tilde{\mathcal{H}}$
\STATE $Side$={\em RND}(\{Left, Right\}) $RSide=$ Reverse of $Side$
\FOR{$l \in [1, N_L]$} 
\STATE $\tilde{\mathcal{H}}=${\em SideRemove}($\tilde{\mathcal{H}}$, $RSide$); $\tilde{\mathcal{H}}$={\em SideAdd}($\tilde{\mathcal{H}}$, $Side$)
\STATE {\bf if}~~$|\tilde{\mathcal{H}}| < k_O$~~{\bf break} ~\COMMENT{Stop search if {\em SideAdd} fails}
\STATE {\bf if}~~$d(\tilde{\mathcal{H}})<d_{N}$~~{\bf then} $d_{N}=d(\tilde{\mathcal{H}})$; $\tilde{\mathcal{H}}_{N}= \tilde{\mathcal{H}}$
\ENDFOR
\STATE {\bf if}~~$d_{N} \le d_O$~~{\bf or}~~$\frac{0.5}{(d_{N}-d_{O})^{\beta}}>${\em RND}(0, 1)~~{\bf return} $\tilde{\mathcal{H}}_{N}$
\RETURN $\tilde{\mathcal{H}}_{O}$ ~\COMMENT{The original $\mathcal{H}$ is unchanged}
\end{algorithmic}
\end{algorithm}

\subsubsection{Insert Moves}

Algorithm \ref{alg:InsertMove} gives the realization of the {\em InsertMove} operator to work on the input $\tilde{\mathcal{H}}$ for obtaining an admissible output.

The operator is realized in three levels, as defined by the parameter $Level \in \{0, 1, 2\}$. For each $v$ in a compact set $\mathcal{V}_{in}=[h_1, h_{|\tilde{\mathcal{H}}|}] \cap \mathcal{V}-\tilde{\mathcal{H}}$ (Line 2), the violation count $\Delta$ is calculated using Algorithm \ref{alg:VioCheck} (Line 3). At level 0, the value $v$ is immediately inserted into $\tilde{\mathcal{H}}$ if $\Delta \equiv 0$ (Line 4). Otherwise, if $Level>0$ and $\Delta \equiv 1$, the violation row $i$ is found using {\em VioRow} (Line 5), which is simply realized by returning $i$ as the conditions are satisfied at Line 3 of Algorithm \ref{alg:VioCheck}, and then $v$ is stored into the set $Q_i$ (Line 5) starting from $\varnothing$ (Line 1). 

At levels 1 and 2, we compare $|Q_i|$ and $m_{i,sb}$, where $sb$ is the second best location in row $i$ of $\mathcal{M}$. Based on Property \ref{thm:occuelems}, $|W_{i,sb}|=m_{i,sb}$. Note that the admissibility is retained after adding elements in $Q_i$ and removing elements in $W_{i,sb}$.

\begin{remark}
\label{thm:insertmoves}
For $\tilde{\mathcal{H}}$={\em InsertMove}($\tilde{\mathcal{H}}$), $d(\tilde{\mathcal{H}})$ is non-increasing at all levels. $|\tilde{\mathcal{H}}|$ is respectively increased by 1 and $|Q_i|-m_{i,sb}$ at levels 0 and 1, and keeps unchanged at level 2. 
\end{remark}

In general, {\em InsertMove} is successful if it can increase $|\tilde{\mathcal{H}}|$. However, the neighborhood might contains too many infeasible moves, as many 1-moves would trigger multiple violations. It might be inefficient to use systematic {\em adjustments} \cite{polymath2014as}. Our implementation targets on feasible moves intelligently by utilizing the violation check clues. 

\subsubsection{Local Search}

Algorithm \ref{alg:LocalSearch} gives the realization of the {\em LocalSearch} operator. We will only focus on the case of improving the input state with $|\tilde{\mathcal{H}}|=k$. Let the input have $d_0=d(\tilde{\mathcal{H}})$. The {\em SideRemove} operator is first applied for $N_S$ times (Lines 1-3). Its output has $|\tilde{\mathcal{H}}|<k$, and $d_1=d(\tilde{\mathcal{H}}) < d_0$.     
The {\em InsertMove} operator is then applied for up to $N_I$ times (Lines 4-6). Based on Remark \ref{thm:subproblem}, the output has $d_2=d(\tilde{\mathcal{H}}) \le d_1$. If this step leads to $|\tilde{\mathcal{H}}| \ge k$, the final output after repairing definitely has a lower diameter than $d_0$. Otherwise, it is still possible to produce a better output as the state is being repaired (Line 7).

\subsection{Region-based Adaptive Local Search (RALS)}

The region-based adaptive local search (RALS) is realized to tackle the problem decomposition as described in Remark \ref{thm:subproblem}. 

Let us consider the problem along the dimension of the numbers in $\mathcal{V}$. For each $\tilde{\mathcal{H}}_k$, it can be indexed by $(h_1, d(\tilde{\mathcal{H}}_k))$. Let $f^*({v})$  be the optimal diameter for all $\tilde{\mathcal{H}}_k$ with $h_1=v$, we can form a set of points $\{(v_1, f^*(v_1)), \cdots, (v_{|\mathcal{V}|}, f^*(v_{|\mathcal{V}|})\}$. It can be seen as a one-dimensional fitness landscape representing the fitness function $f^*(v)$ on the discrete variable from $v\in \mathcal{V}$. Note that the optimal solution on this fitness landscape is the optimal solution of the original problem. 

Essentially, we would like to focus the search on those promising regions where $f^*(v)$ has higher quality. Nevertheless, early search can provide some clues for narrowing down promising regions, even though the fitness landscape itself is not explicit at the beginning, as $f^*(v)$ at each $v$ can be revealed through extensive local search.  

\begin{algorithm} [t]                   
\caption{{\em InsertMove}: Local moves in $[h_1, h_{|\tilde{\mathcal{H}}|}]$ of $\tilde{\mathcal{H}}$}   
\label{alg:InsertMove}                          
\begin{algorithmic}[1]                    
\REQUIRE $\tilde{\mathcal{H}}$ \COMMENT{Parameter: $Level \in \{0, 1, 2\}$ }
\STATE Initialize $\{Q_i=\varnothing | i\in [1, |\mathcal{P}|]\}$ \COMMENT{Use as $Level > 0$}
\FOR{$v \in \mathcal{V}_{in}=[h_1, h_{|\tilde{\mathcal{H}}|}] \cap \mathcal{V}-\tilde{\mathcal{H}}$}
\STATE $\Delta=${\em VioCheck}($v$,  $\tilde{\mathcal{H}}$)  \COMMENT{Algorithm \ref{alg:VioCheck}}
\STATE {\bf if}~~$\Delta\equiv 0$~~{\bf return}~~$\tilde{\mathcal{H}}=\tilde{\mathcal{H}} \cup \{v\}$  \COMMENT{Level 0: Insert one number}
\STATE {\bf if}~~$\Delta\equiv 1$~~{\bf then}~~$i=\text{\em VioRow}(v, \tilde{\mathcal{H}})$; $Q_i=Q_i \cup \{v\}$ 
\ENDFOR
\FOR{$Level>0$~~{\bf and}~~$i \in [1, |\mathcal{P}|]$}
\STATE {\bf if}~~$|Q_i|>m_{i,sb}$~~{\bf return}~~$\tilde{\mathcal{H}}=\tilde{\mathcal{H}}+Q_i-W_{i, sb}$ \COMMENT{Level 1}
\ENDFOR
\FOR{$Level>1$~~{\bf and}~~$i \in [1, |\mathcal{P}|]$ (In Random Order)}
\STATE {\bf if}~~$|Q_i| \equiv m_{i,sb}>0$~~{\bf return}~~$\tilde{\mathcal{H}}=\tilde{\mathcal{H}}+Q_i-W_{i, sb}$ 
\ENDFOR
\RETURN $\tilde{\mathcal{H}}$ ~\COMMENT{The original $\mathcal{H}$ is unchanged}
\end{algorithmic}
\end{algorithm}

\begin{algorithm} [t]                     
\caption{{\em LocalSearch}: Remove \& insert to improve $\tilde{\mathcal{H}}_k$} 
\label{alg:LocalSearch}                           
\begin{algorithmic}[1]                   
\REQUIRE $\tilde{\mathcal{H}}$, $N_S$, $N_I$ ~\COMMENT{Parameters: $N_S \ge 1$, $N_I \ge 1$}
\FOR{$n\in[1, N_S]$}
\STATE $Side$={\em RND}(\{Left,Right\}); $\tilde{\mathcal{H}}$={\em SideRemove}($\tilde{\mathcal{H}}$, $Side$)
\ENDFOR
\FOR{$n\in[1, N_I]$}
\STATE $\tilde{\mathcal{H}}=${\em InsertMove}($\tilde{\mathcal{H}}$);~~{\bf if}~~$|\tilde{\mathcal{H}}|\ge k$~~{\bf break}
\ENDFOR
\RETURN $\tilde{\mathcal{H}}_k=${\em Repair}($\tilde{\mathcal{H}}$) 
\end{algorithmic}
\end{algorithm}

\subsubsection{Database Management}

We use a simple database, denoted by \text{DB}, to keep the high-quality solutions $\tilde{\mathcal{H}}_k$ found during the search, and index each of them as $(v, f({v}))$, where $v=h_1$, and $f({v})=d(\tilde{\mathcal{H}}_k)$ for each $\tilde{\mathcal{H}}_k$. For each $v$, only the best-so-far solution and the corresponding $f({v})$ is kept. Here $f({v})$ plays the role of a virtual fitness function that is updated during the search process to approximate the real fitness function $f^*(v)$. 

The database is managed in a region-based mode. Specifically, the total range $[0, U]$ of the numbers in $\mathcal{V}$ is divided into $N_R$ regions. There are three basic operations for DB. 

The {\em DBInit} operator is used for providing the initialization. The greedy sieve \cite{polymath2014variants} is applied to generate a state $\tilde{\mathcal{H}}_k$ in each region for forming the initial $f({v})$. 

The {\em DBSelect} operator is used for selecting one incumbent state to apply the search operation. In the region-based mode, there are two steps to provide the selection. In the first step, each region provides one candidate. In this paper, we greedily choose the best solution in each region. In the second step, the incumbent state is selected from the candidates provided by all regions. We consider the following implementation. At the probability $\gamma$, the candidate is selected at random. Otherwise, {\em tournament selection} is applied to select the best solution among totally $N_T$ randomly chosen candidates.

The {\em DBSave} operator simply stores each $\tilde{\mathcal{H}}_k$ into DB, and updates $f({v})$ internally. Dominated solutions are discarded.

\subsubsection{Algorithm Realization}

Algorithm \ref{alg:RLAS} gives the implementation of RALS to obtain $\mathcal{H}_k^*$ for a given $k$. First, $\mathcal{V}$ and $\mathcal{P}$ are initialized using Algorithm \ref{alg:SieveVP}, using $U=\lceil 1.5\cdot (k\log k + k) \rceil$ for the range $[0, U]$ (Line 1) to ensure $U>d_k^{UB}$. Afterward, the database DB with $N_R$ regions is initiated using the {\em DBInit} operator (Line 2). 

The search process runs $T$ iterations in total. In each iteration, we first select one incumbent solution $\tilde{\mathcal{H}}_k$ from DB using the {\em DBSelect} operator (Line 4). Then the actual search tackles two parts of the problem. The {\em ShiftSearch} operator is used to search on the virtual fitness landscape $f({v})$ (Line 5).  
The {\em LocalSearch} operator is then applied to improve $f({v})$ locally (Lines 6-7). For each search operator in Lines 5-7, the {\em DBSave} operator is applied to store newly generated solutions. Finally, the best solution $\mathcal{H}^*$ in DB is returned.

\begin{algorithm} [t]      
\caption{RLAS algorithm to obtain $\mathcal{H}_k^*$ for a given $k$}   
\label{alg:RLAS}                          
\begin{algorithmic}[1]                   
\STATE Intialize $\mathcal{V}$, $\mathcal{P}$ using Algorithm \ref{alg:SieveVP} \COMMENT{$U=1.5 \cdot \lceil k\log k + k \rceil$}
\STATE DB={\em DBInit}($N_R$) \COMMENT{Initiate DB with $N_R$ regions} 
\FOR{$t \in [1, T]$}
\STATE $\tilde{\mathcal{H}}_k=${\em DBSelect}(DB)~\COMMENT{Select one incumbent solution from DB}
\STATE $\tilde{\mathcal{H}}_k=${\em ShiftSearch}($\tilde{\mathcal{H}}_k$); {\em DBSave}($\tilde{\mathcal{H}}_k$, DB)
\STATE $\tilde{\mathcal{H}}_k=${\em LocalSearch}($\tilde{\mathcal{H}}_k$, 1, $N_{I1}$); {\em DBSave}($\tilde{\mathcal{H}}_k$, DB)
\STATE $\tilde{\mathcal{H}}_k=${\em LocalSearch}($\tilde{\mathcal{H}}_k$, 2, $N_{I2}$); {\em DBSave}($\tilde{\mathcal{H}}_k$, DB)
\ENDFOR
\RETURN $\mathcal{H}^*$ in DB~\COMMENT{Return the best solution stored in DB}
\end{algorithmic}
\end{algorithm}

\section{Results and Discussion} \label{sec:results}

We now turn to the empirical evaluation of the proposed algorithm. For the benchmark instances, we refer to an online database \cite{Sutherl2015b} that has been established and extensively updated to contain the narrowest admissible $k$-tuples known for all $k\le 5000$. The algorithm is coded in Java, and our experiments were run on AMD 4.0GHz CPU. For each instance, 100 independent runs were performed.

\subsection{Results by Existing Methods}

Most existing techniques to solve this problem are constructive and sieve methods \cite{polymath2014as}. The sieve methods are realized by sieving an integer interval of residue classes modulo primes $p < k$ and then selecting an admissible $k$-tuple from the survivors. The easiest way to construct a narrow $\tilde{\mathcal{H}}_k$ is using the first $k$ primes past $k$ \cite{zhang2014bounded}. As an optimization, the sieve of Eratosthenes takes $k$ consecutive primes, to search starting from $p<k$, in order to select one among the admissible tuples that minimize the diameter. 

The Hensley-Richards sieve \cite{hensley1974primes} uses a heuristic algorithm to sieve the interval $[-x/2,x/2]$ to obtain $\tilde{\mathcal{H}}_k$, leading to the upper bound \cite{polymath2014variants}: 
\[
H(k)\le k\log k + k\log\log k -(1+\log 2)k + o(k).
\]

The Schinzel sieve, as also considered in \cite{gordon1998dense,clark2001dense}, sieves the odd rather than even numbers. In the shifted version \cite{polymath2014variants}, it sieves an interval $[s,s+x]$ of odd integers and multiples of odd primes $p\le p_m$, where $x$ is sufficient large to ensure at least $k$ survivors, and $m$ is sufficient large to ensure that the survivors form $\tilde{\mathcal{H}}$, $s\in [-x/2,x/2]$ is the starting point to choose for yielding the smallest final diameter.

As a further optimization, the shifted greedy sieve \cite{polymath2014variants} begins as in the shifted Schinzel sieve, but the minimally occupied residue class are greedily chosen to sieve for primes $p > \tau \sqrt{k\log k}$, where $\tau$ is a constant. Empirically, it appears to achieve the bound \cite{polymath2014as}:
\[
H(k) \le k\log k + k + o(1).
\]

Table~\ref{tab:extsol} lists the upper bounds obtained by applying a set of existing techniques, including $k$ primes past $k$, Eratosthenes (Zhang) sieve, Hensley-Richards sieve, Schinzel and Shifted Schinzel sieve, by running the code\footnote{\url{http://math.mit.edu/~drew/ompadm_v0.5.tar}} provided in \cite{polymath2014variants}, on $k=\{1000, 2000, 3000, 4000, 5000\}$. The best known results are retrieved from  \cite{Sutherl2015b}. 

\begin{table}
\centering
\caption{Upper bounds on $\mathcal{H}_k$ by existing sieve methods.}\label{tab:extsol}
\includegraphics[width=0.43\textwidth]{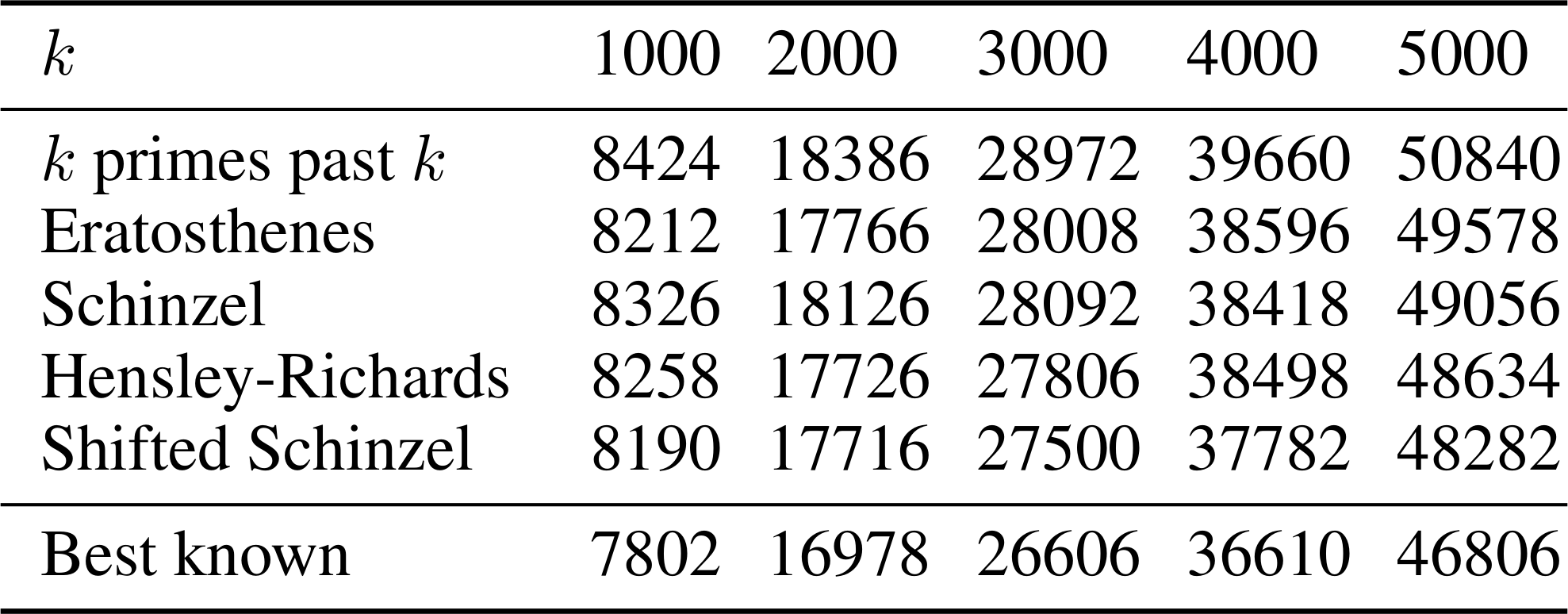}
\end{table}

\begin{table}
\centering
\caption{Upper bounds on $\mathcal{H}_k$ by different RALS versions.}\label{tab:testsol}
\includegraphics[width=0.43\textwidth]{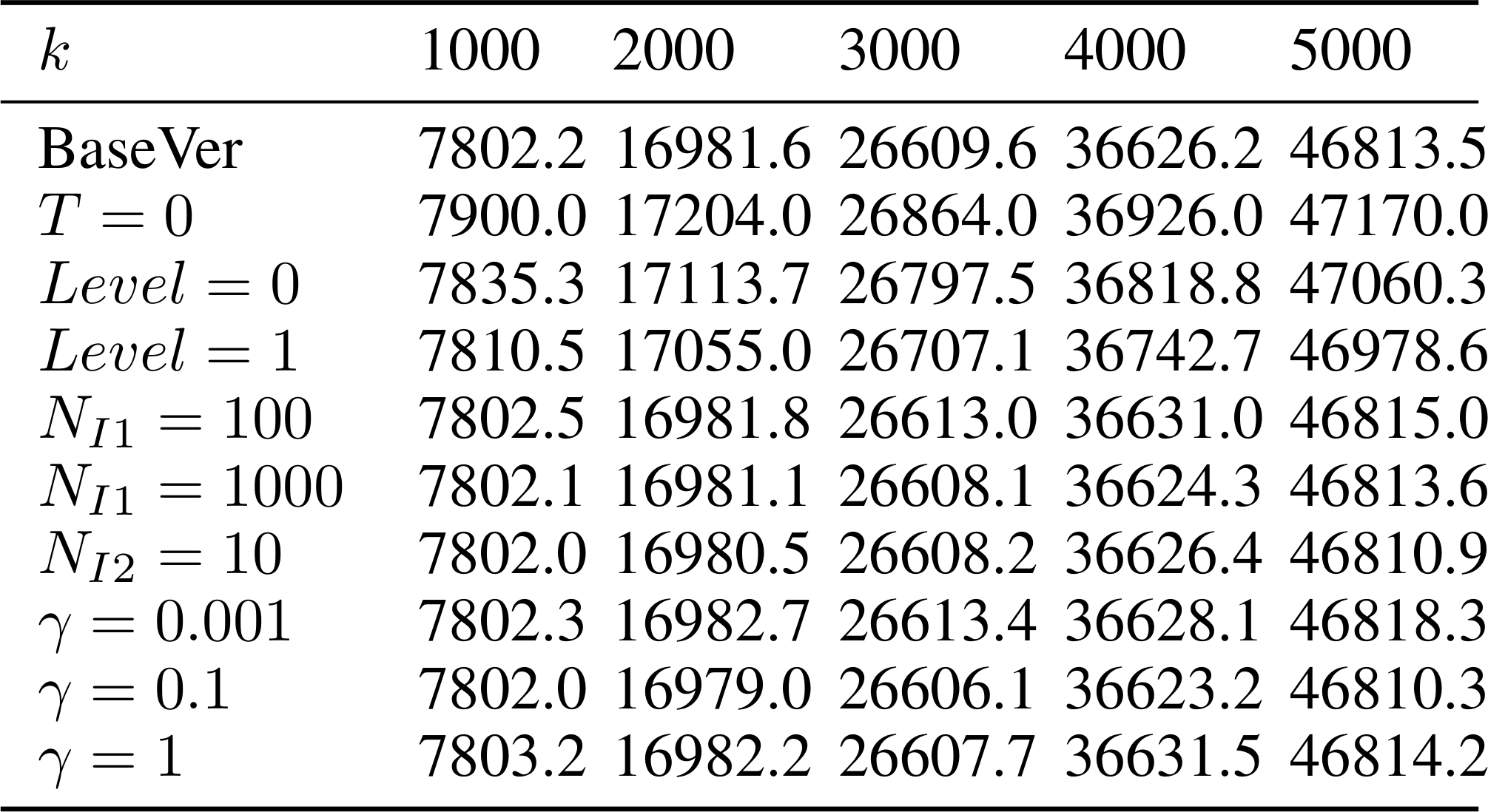}
\end{table}

\begin{figure*} [t]
\centering
\begin{subfigure}{.195\textwidth}
  \centering
  \includegraphics[width=.95\linewidth]{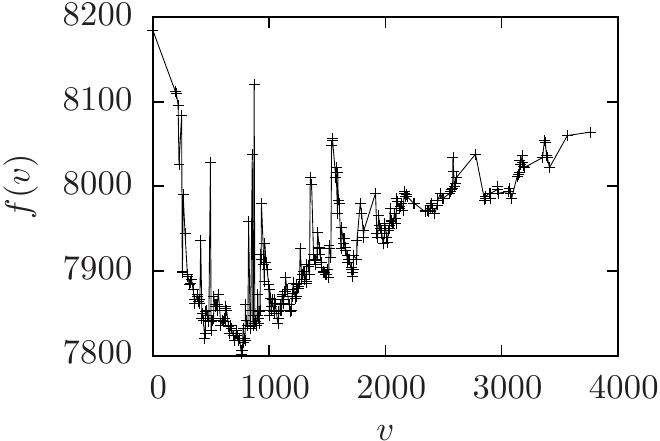}
  \caption{$k=1000$}
  \label{fig:fitness1K}
\end{subfigure}%
\begin{subfigure}{.195\textwidth}
  \centering
  \includegraphics[width=.95\linewidth]{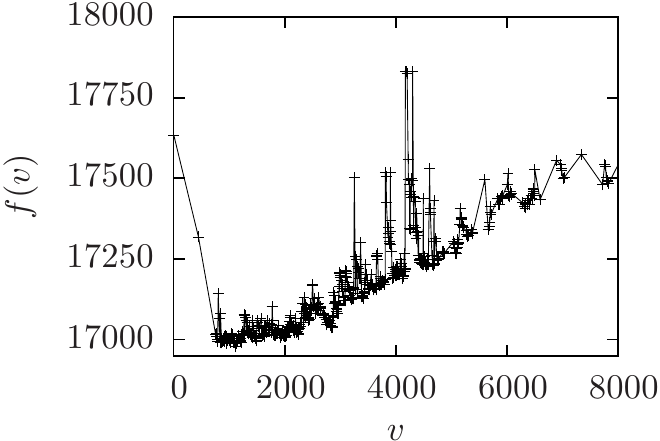}
  \caption{$k=2000$}
  \label{fig:fitness2K}
\end{subfigure}
\begin{subfigure}{.195\textwidth}
  \centering
  \includegraphics[width=.95\linewidth]{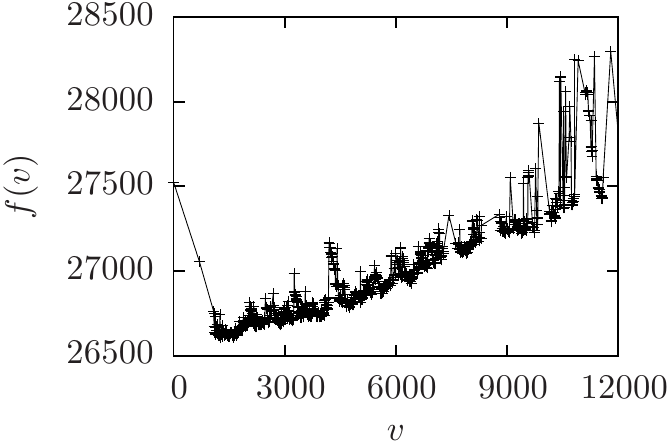}
  \caption{$k=3000$}
  \label{fig:fitness3K}
\end{subfigure}
\begin{subfigure}{.195\textwidth}
  \centering
  \includegraphics[width=.95\linewidth]{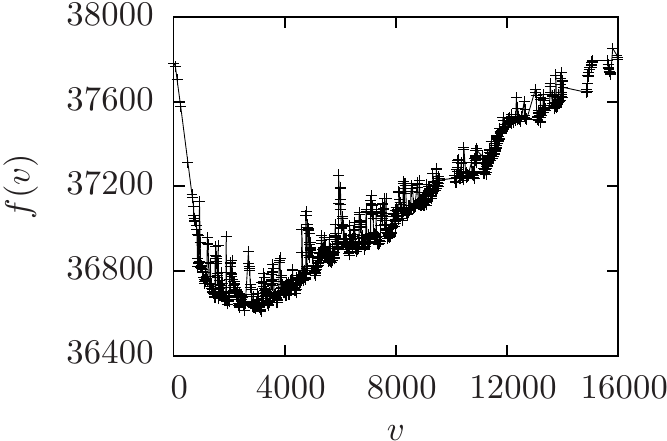}
  \caption{$k=4000$}
  \label{fig:fitness4K}
\end{subfigure}
\begin{subfigure}{.195\textwidth}
  \centering
  \includegraphics[width=.95\linewidth]{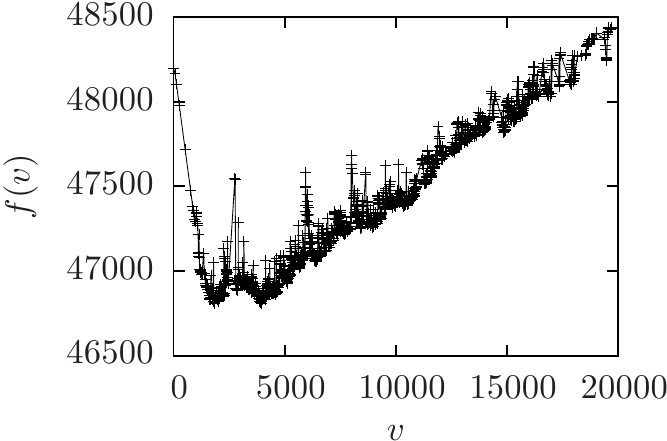}
  \caption{$k=5000$}
  \label{fig:fitness5K}
\end{subfigure}
\caption{Snapshot of the virtual fitness landscape $f(v)$, taking $v$ as the start element of admissible $k$-tuples.}
\label{fig:SnapshotVirtual}
\end{figure*}

\subsection{Results by RALS algorithm}

Table ~\ref{tab:testsol} lists the average results of different versions of the proposed RALS algorithm. The ``BaseVer'' version is defined with the following settings. For the database DB, we use $N_R=20$. For its {\em DBSelect} operator, there are $\gamma=0.01$ and $N_T=4$. For the search loop, we consider $T=1000$ iterations. For the {\em ShiftSearch} operator, we set $N_L=10$ and ${\beta}=1$. For the {\em InsertMove} operator, there is $Level=2$. For the parameters of {\em LocalSearch} in Algorithm \ref{alg:RLAS}, we set $N_{I1}=500$ and $N_{I2}=0$. The other versions are then simply the ``BaseVer'' version with different parameters.

With $T=0$, the algorithm returns the best results obtained by the shifted greedy sieve \cite{polymath2014as} in the $N_R$ regions. The results are significantly better than the sieve methods in Table~\ref{tab:extsol}. The search operators in RALS show their effectiveness as all RALS versions with $T>0$ perform significantly better than the version with $T=0$. 

Note that ``BaseVer'' has $Level=2$, we can compare the RALS versions with different levels $\{0, 1, 2\}$ in the {\em InsertMove} operator of {\em LocalSearch}. On the performance, the version with a higher level produces better results than that of a lower level. With greedy search only, the first {\em LocalSearch} works as an efficient {\em contraction} process \cite{polymath2014as}. As described in Remark \ref{thm:insertmoves}, {\em InsertMove} performs greedy search at levels 0 and 1, but performs plateau moves at level 2, from the perspective of updating $|\tilde{\mathcal{H}}|$. At level 0, the search performs elemental 1-moves. At level 1, the search can be in a very large neighborhood although it has a low time complexity. Plateau moves is used at level 2 to find exits, as remaining feasible moves are more difficult to check. Finding exits to leave plateaus \cite{hoffmann2001local,frank1997gravity} has been an important research topic about the local search topology on many combinatorial problems \cite{bonet2001planning,benton2010g,sutton2010directed}. From the viewpoint of the {\em LocalSearch} operator, the plateau moves on the part solved by {\em InsertMove} help escaping from local mimina in the landscape of the subproblem. 

\begin{table}
\centering
\caption{Results of ``BaseVer'' with $\gamma=0.1, N_{I2}=10$.}\label{tab:bestConf}
\includegraphics[width=0.44\textwidth]{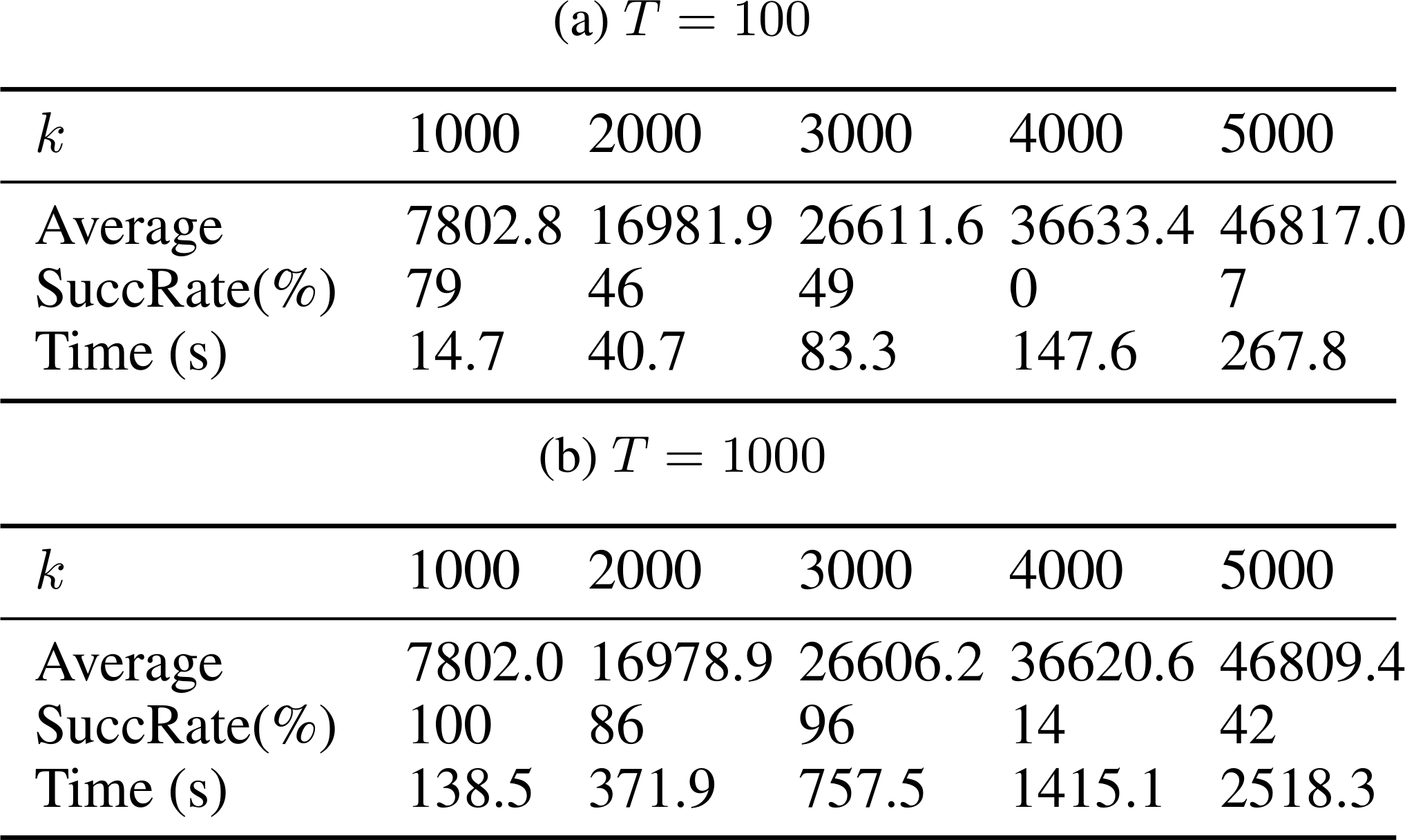}
\end{table}

We compare the RALS versions with different $N_{I1}\in\{100, 500, 1000\}$, as ``BaseVer'' has $N_{I1}=500$. Especially for $k \in \{3000, 4000\}$, the improvements of using higher $N_{I1}$ are extremely significant as $N_{I1}$ increases from 100 to 500, but are less significant as $N_{I1}$ further increases to 1000. 

In ``BaseVer'', the second {\em LocalSearch} in Line 7 of Algorithm \ref{alg:RLAS} is actually not used if it has $N_{I2}=0$. As we increase $N_{I2}$ to $10$, the instance $k=1000$ can be fully solved to the best known solution, and the instance $k=5000$ can also be solved to obtain a significantly better result. 

Lines 1-3 of Algorithm \ref{alg:LocalSearch} might be viewed as perturbation, an effective operator in stochastic local search \cite{hoos2004stochastic} to escape from local minima on rugged landscapes \cite{tayarani2014landscape,billinger2013search}. In RALS, the second {\em LocalSearch} essentially applies a larger perturbation than the first {\em LocalSearch}.   

Table \ref{tab:testsol} also gives the comparison for RALS with $\gamma \in \{0.001, 0.01, 0.1, 1\}$ for selecting the incumbent state in {\em DBSelect}. The larger the $\gamma$, the more random the selection is. The best performance is achieved as $\gamma=0.1$, neither too greedy nor too random. We can gain some insights from a typical snapshot of the virtual fitness landscape $f(v)$, as shown in Figure \ref{fig:SnapshotVirtual}. It is easy to spot the valley with high-quality solutions, as they provide significant clues for adaptive search. 
Meanwhile, the noises on the fitness landscape might reduce the effectiveness of pure greedy search. Thus, there is a trade-off between greedy and random search.

\begin{table}
\centering
\caption{New upper bounds on $\mathcal{H}_k$ for $k \in [2500, 5000]$.}\label{tab:newsol}
\includegraphics[width=0.424\textwidth]{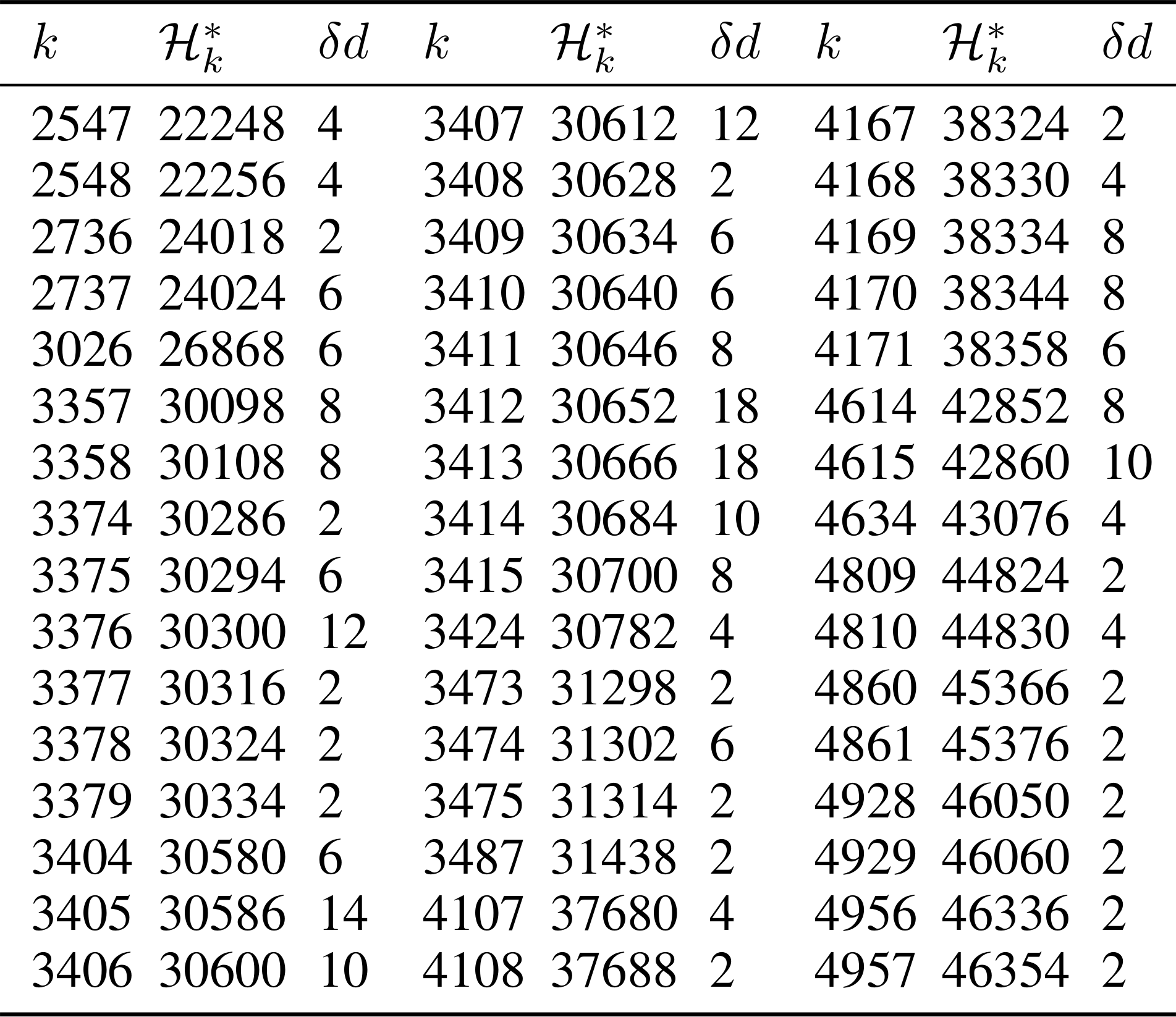}
\end{table}

Tables \ref{tab:bestConf} lists the performance measures, including the average, the successful rate of finding best known solutions (SuccRate), and the calculating time,  by the ``BaseVer'' version with both $\gamma=0.1$ and $N_{I2}=10$, as $T=100$ and $T=1000$. This version achieves high SuccRate for $k\in\{1000, 2000, 3000\}$, and moderate SuccRate for $k\in\{4000, 5000\}$, as $T=1000$. It also reaches reasonable good SuccRate as $T=100$, with a lower execution time. 

Finally, we apply RLAS to compare the results for $k \in [2500,5000]$ in \cite{Sutherl2015b}. In Table \ref{tab:newsol}, we list the new upper bound $\mathcal{H}_k^*$ and the improvement on the diameter $\delta d$ for each $k$ of the 48 instances. Eight instances among them have $\delta d \ge 10$. Thus, AI-based methods might make further contributions to pure mathematics applications.

\section{Conclusions} \label{sec:Conclusion}

We presented a region-based adaptive local search (RALS) method to solve a case of pure mathematics applications for finding narrow admissible tuples. We formulated the original problem into a combinatorial optimization problem. We showed how to exploit the local search structure to tackle the combinatorial landscape, and then to realize search strategies for adaptive search and for effective approaching to high-quality solutions. Experimental results demonstrated that the method can efficiently find best known and new solutions.

There are several aspects of this work that warrant further study. A deeper analysis might be applied to better identify properties of the local search topology on the landscape. One might also apply advanced AI strategies, e.g., algorithm portfolios \cite{gomes2001algorithm} and SMAC \cite{hutter2011sequential}, to obtain an even greater computational advantage.

\newpage 

\bibliographystyle{named}
\bibliography{ijcai16}

\end{document}